\documentclass[sigconf,nonacm]{acmart}
\AtBeginDocument{%
  \providecommand\BibTeX{{%
    \normalfont B\kern-0.5em{\scshape i\kern-0.25em b}\kern-0.8em\TeX}}}

\usepackage{tikz}
\usetikzlibrary{shapes, decorations, arrows, calc, arrows.meta, fit, positioning}
\tikzset{
    -Latex,auto,node distance =1 cm and 1 cm,semithick,
    state/.style ={ellipse, draw, minimum width = 0.7 cm},
    point/.style = {circle, draw, inner sep=0.04cm, fill},
    bidirected/.style={Latex-Latex,dashed},
    el/.style = {inner sep=2pt, align=left, sloped}
}

\usepackage{bm}
\usepackage{comment}
\usepackage{caption}
\usepackage{subcaption}
\usepackage{booktabs, multirow, makecell, array}
\newcommand{\ra}[1]{\renewcommand{\arraystretch}{#1}} 

\hypersetup{
    breaklinks=true,
    urlcolor  = black,
	colorlinks = true,
	citecolor = blue, 
	linkcolor = blue 
}

\renewcommand{\b}{\beta}
\newcommand{\indep}{\perp\!\!\!\perp}

\begin{document}

\title[Investigating Bias with a Synthetic Data Generator]{Investigating Bias with a Synthetic Data Generator: Empirical Evidence and Philosophical Interpretation}


\author{Alessandro Castelnovo}
\email{a.castelnovo5@campus.unimib.it}
\email{alessandro.castelnovo@intesasanpaolo.com}

\affiliation{%
  \institution{Intesa Sanpaolo S.p.A., \\ Dept. of Informatics, Systems \& Communication, Univ. of Milan-Bicocca}
  \city{Milan}
  \country{Italy}
}

\author{Riccardo Crupi}
\email{riccardo.crupi@intesasanpaolo.com}

\author{Nicole Inverardi}
\email{nicole.inverardi@intesasanpaolo.com}

\author{Daniele Regoli}
\email{daniele.regoli@intesasanpaolo.com}
\orcid{0000-0003-2711-8343}

\author{Andrea Cosentini}
\email{andrea.cosentini@intesasanpaolo.com}

\affiliation{%
  \institution{Intesa Sanpaolo S.p.A.}
  \streetaddress{Piazza Paolo Ferrari, 10}
  \city{Milan}
  \country{Italy}
}

\renewcommand{\shortauthors}{Castelnovo, et al.}

\begin{abstract}
Machine learning applications are becoming increasingly pervasive in our society. Since these decision-making systems rely on data-driven learning, risk is that they will systematically spread the bias embedded in data. In this paper, we propose to analyze biases by introducing a framework for generating synthetic data with specific types of bias and their combinations. We delve into the nature of these biases discussing their relationship to moral and justice frameworks. Finally, we exploit our proposed synthetic data generator to perform experiments on different scenarios, with various bias combinations. We thus analyze the impact of biases on performance and fairness metrics both in non-mitigated and mitigated machine learning models. 
\end{abstract}

\keywords{Synthetic Data, Bias, Fairness, Worldview, Machine Learning}

\maketitle

\section{Introduction}

As society grows more digital, a greater amount of data becomes accessible for decision-making. In this context, machine learning techniques are increasingly being adopted by businesses, governments, and organizations in many important domains that affect people’s lives every day. However, algorithms, like humans, are susceptible to biases that might lead to unfair outcomes \cite{angwin2016machine}.

Bias is not a recent problem; rather, it is ingrained in human society and, as a result, it is reflected in data \cite{ntoutsi2020bias}. The risk is that the adoption of machine learning algorithms could amplify or introduce biases that will reoccur in society in a perpetual cycle \cite{mehrabi2021survey}.

Several projects and initiatives have been launched in recent years aimed at bias mitigation and the development of fairness-aware machine learning models. Following to \cite{ntoutsi2020bias}, we divide these works in three main categories:

\begin{itemize}
    \item \textit{Understaning bias}. Approaches that help to understand how bias is generated in society and manifest in data. This category contains studies of differences among biases, as well as their definition and formalization.
    \item \textit{Accounting for bias}. Approaches discussing how to manage bias depending on the context, the regulation, the vision and the strategy on fairness \cite{castelnovo2021towards, castelnovo2020befair}. As discussed in \cite{friedler2021possibility}, different definitions of fairness and their implementations correspond to different axiomatic beliefs about the world (or  \textit{worldviews}), in general mutually incompatible.
    \item \textit{Mitigating bias}. Technical approaches aimed to the development of machine learning models devoted to bias reduction with performance optimization. Depending on the stage of the machine learning pipeline where the bias is mitigated, these methods are typically divided into: \textit{pre-processing}~\cite{madaio2020co,kamiran2009classifying}, \textit{in-processing}~\cite{agarwal2018reductions} and \textit{post-processing}~\cite{hardt2016equality,lohia2019bias}.

\end{itemize}

One common approach to investigate the nature of bias is to conduct experiments on \emph{ad hoc} scenarios through the generation of synthetic data~\cite{le2022survey,raghunathan2021synthetic}. The benefits of this strategy include the possibility of examining circumstances not available with real-world data but that may occur, and ---even when real-world data is available--- to precisely control and understand the data generation mechanism. Moreover, it is indisputable that making data, and related challenges, accessible to the research community for analysis could be of help for the development of sound policy decisions and benefit society~\cite{raghunathan2021synthetic}.

\subsection{Contribution}

With this work, we aim to contribute to the literature on fairness in machine learning in each of the three areas discussed above through the use of synthetic data. 

In particular, we contribute to \textit{understanding bias} by introducing a model framework for generating synthetic data with specific types of bias. Our formalisation of these various types of bias is based on the theoretical classifications present in the relevant literature, such as the surveys on bias in machine learning by \citeauthor{mehrabi2021survey} and \citeauthor{ntoutsi2020bias}. 

Against the background of the stream of literature about the relation between moral worldviews and biases, and in particular following \cite{friedler2021possibility, hertweck2021moral, heidari2019moral}, we analyze the worldviews related with each bias that our framework is able to generate, thus providing some insights in the discussion on \textit{accounting for bias}.

Finally, about \textit{mitigating bias}, we leverage our framework to generate twenty-five different scenarios characterized by the presence of various bias combinations. In each setting, we investigate the behavior and effects of traditional machine learning mitigation strategies~\cite{bird2020fairlearn}.

An open source implementation of the proposed model framework is available at \href{https://github.com/rcrupiISP/ISParity}{\texttt{github.com/rcrupiISP/ISParity}}. 

\section{Related Works\label{sec:related_work}}

\subsection{Synthetic data}
Synthetic data generation is a relevant practice for both businesses and the scientific community. As a result, the literature has given it a lot of attention. 

Main direction behind the generation of synthetic data are: the \textit{emulation} of certain key information in real dataset while preserving privacy \cite{raghunathan2021synthetic, assefa2020generating} and; the \textit{generation} of different test scenario for evaluating phenomenon not covered by the available data \cite{le2022survey}.
\citeauthor{assefa2020generating}\cite{assefa2020generating} presented basic use cases with specific examples in the financial domain like: internal data use restrictions, data sharing, tackling class imbalance, lack of historical data and training advanced Machine Learning models. Moreover, the authors defined \textit{Privacy preserving}, \textit{Human readability} and \textit{Compactness} as desirable properties for synthetic representation.

Is important to remark that synthetic data generation is not a possible technique for data anonymisation \cite{majeed2020anonymization}, but an alternative data sanitisation method to data masking for preserving privacy in published data \cite{surendra2017review}. In fact, synthetic data are typically randomly generated with constraints to protect sensitive private information and retain valid inferences with the attributes in the original data \cite{raghunathan2021synthetic}. The provided synthetic data are generally
classified into: \textit{fully synthetic data}, \textit{partially synthetic data} and \textit{hybrid synthetic data} (see \cite{aggarwal2008privacy} for further details). We refer to \cite{surendra2017review} for a detailed overview of the techniques for generating synthetic data sets.

As introduced in the previous section, we are interested in produce fully synthetic data to replicate some common biases that could affect the data, and investigates fairness-related issues that arise from the development of machine learning models on them. In this regard, there are numerous works in the literature that generate synthetic datasets to simulate desired scenarios and, from these, testing discrimination-aware methods \cite{d2020fairness, loh2019subgroup, reddy2021benchmarking, castelnovo2022clarification}.
For examples, \citeauthor{reddy2021benchmarking} evaluate different fairness methods trained with deep neural networks on synthetic dataset. In these data were present different imbalanced and correlated data configuration to verify the limits of the current models and better understand in which setups they are subject to failure.

We contribute in this field of literature by introducing a modeling framework to generate synthetic data presenting specific forms of bias. Our formalisation of these different kinds of bias builds upon theoretical classifications present in the relevant literature, such as the works of \citeauthor{mehrabi2021survey} and \citeauthor{ntoutsi2020bias}. We leverage our proposed method to actually generate several datasets, each characterized by a specific combination of biases, and perform experiments on them to examine the effects of such biases on state-of-the-art mitigation approaches.

\subsection{Bias and moral framework in decision-making}

There is little consensus in the literature regarding bias classification and taxonomy. 
Moreover, the very notion of bias depends on profound ethical and philosophical considerations, which is likely one of the very causes for the lack of consensus. Different understandings of bias and fairness depend on the assumption of a belief system beforehand. \citeauthor{friedler2021possibility} and \cite{hertweck2021moral} talks about \textit{worldviews}. In particular, \cite{friedler2021possibility} outlines two extreme cases, referred to as \textit{What You See Is What You Get} (WYSIWYG) and \textit{We are All Equal} (WAE).



Starting from the definition of three different metric spaces, these two perspectives differ because of the way they consider the relations in between. The first space is named \textit{Construct Space} (CS) and represents all the unobservable realized characteristics of an individual, such as intelligence, skills, determination or commitment. The second space is the \textit{Observable Space} (OS) and contains all the measurable properties that aim to quantify the unobservable features, think e.g. of IQ or aptitude tests. The last space is the \textit{Decision Space} (DS), representing the set of choices made by the algorithm on the basis of the measurements available in OS. Note that shades of ambiguity are already detectable at this level, because the mappings between spaces are susceptible of distortions. Moreover, CS is by definition unobservable, thus we can only make \emph{assumptions} on it.

According to WYSIWYG, CS and OS are essentially equal and any distortion between the two is altogether irrelevant for the fairness of the decision resulting in DS. Contrarily, WAE don't make assumptions about the similarity of OS and CS, and moreover assumes that we are all equal in CS, i.e. that any difference between CS and OS is due to a biased observation process that results in an unfair mapping between CS and OS. With the distance between worldviews in mind, the notion of fairness inspired by~\cite{dwork2012fairness} affirming that individuals that are close in CS shall also be close in DS (commonly known as individual fairness) appears diversified and differently achievable. If WYSIWYG is chosen, non-discrimination is guaranteed as soon as the mapping between OS and DS is fair, since CS  $\approx$ OS. On the other hand, according to WAE the mapping between CS and OS is distorted by some bias whenever a difference among individuals emerges (\cite{hertweck2021moral} call this difference Measurement Bias); therefore, to obtain a fair mapping between CS and DS those biases should be mitigated properly. 

Building on \cite{friedler2021possibility}, \cite{hertweck2021moral} describes a more realistic and nuanced scenario by introducing the notion of \emph{Potential Space} (PS): individuals belonging to different groups may indeed have different realized talents (i.e. they actually differ in CS), and these may be accurately measured by resumes (i.e. CS $\approx$ OS), but, if we assume that these groups have the same \emph{potential} talents (i.e. they are equal in PS), then the realized difference must be due to some form of unfair treatment of one group, that is referred to as \emph{life bias}. \cite{hertweck2021moral} call this view \textit{We Are All Equal in Potential Space} (WAEPS). Actually, as argued in \cite{hertweck2021moral}, we can effectively think of the WAE assumption as a family of assumptions, depending on the point time in which the equaility is assumed to hold: the more we go back in time in assuming equality between individuals, the more the consideration of life circumstances becomes strong, and thus the less discrimination between individuals is considered legitimate.

Given that those extreme worldviews involve heavy assumptions, philosophical theories around \textit{Equality of Opportunity} (EO) offer some suggestions and interpretive tools for approaching biases in different situations~\cite{fleurbaey2008fairness, roemer2015equality, rawls2004theory}. In this sense, western political philosophy and algorithmic fairness literature encounter themselves in the formulation of fairness around the concept of equal opportunities for all members of society. 
\citeauthor{heidari2019moral} list three different EO conceptions, going from more permissive to more stringent: \textit{Libertarian~EO}, by which individuals are held accountable for any characterizing feature, sensitive one included; \textit{Formal~EO}, by which individuals are not held accountable for differences in sensitive features only; \textit{Substantive~EO}, by which there is a set of individual characteristics which are due to \textit{circumstances} and others that are a consequence of individual \textit{effort}, and thus people should be held accountable only on the basis of the latter.  
The choice of which characteristics fall in the level of circumstances and which can be considered as individual effort is far from obvious, and is essentially connected to the point in time in which the WAE assumption holds.

In the following section, we shall describe the most common biases, explaining how they relate to these fundamental concepts in our view.

\section{Fundamental types of bias\label{sec:bias_list}}

Considering that the assumptions about the worldview and the EO framework affect the conception as well as the assessment of biases, in what follows we focus on what we consider the fundamental building blocks of most types of biases, namely: \textit{Historical bias}, \textit{Measurement bias}, \textit{Representation bias}, \textit{Omitted variable bias}.

\textbf{Historical bias} ---sometimes referred to as \textit{social bias}, \textit{life bias}, or \textit{structural bias}~\cite{mehrabi2021survey, ntoutsi2020bias, hertweck2021moral}--- occurs whenever a variable of the dataset relevant to some specific goal or task is dependent on some sensitive characteristic of individuals, \emph{but in principle it should not}. An example of this bias is the different average income among men and women, which is due to long-lasting social pressures in a man-centered society, and does not reflect intrinsic differences among sexes. 
Following~\cite{mehrabi2021survey}, we can talk of a form of bias going \textit{from users to data}: this type of bias affects directly the actual phenomenon generating the data. A similar situation may arise when the dependence of sensitive individual characteristics is present with respect to the variable that we are trying to estimate or predict. For instance, there are cases in which the target of model estimation is itself prone to some form of bias, e.g. because it is the outcome of some human decision. Think e.g. to trying to build a data-driven decision process to decide whether to grant or not a loan on the basis of past loan officers' decisions, and not on actual repayments. Note that the actual presence of historical bias is conditioned by the previous assumption of the WAE worldview. Indeed, arguing that in principle there should be no dependence on some sensitive features makes sense only if a moral belief of substantial equity is required to begin with. Otherwise, according to WYSIWYG, CS is fairly reported in OS and therefore  structural differences between individuals are legitimate sources of inequality. Moreover, accepting the Libertarian EO or the Substantive EO frameworks would involve the legitimate use of some sensitive features, respectively because they are a property of the self and because we should be aware of their influence on the values of non-sensitive features.
Ultimately, the presence of historical bias depends on the assumption of WAE at the initial time of life. As argued in \cite{hertweck2021moral}, interpreting bias as historical means conceiving equality at the level of PS, that describes the innate/native potential of each individual.

\textbf{Measurement bias} occurs when a proxy of some variable relevant to a specific goal or target is employed, and that proxy happens to be dependent on some sensitive characteristics. For instance, one may argue that IQ is not a ``fair'' approximation of actual ``intelligence'', and it might systematically favour/disfavour specific groups of individuals. Statistically speaking, this type of bias is not very different from historical bias ---since it results in employing a variable correlated with sensitive attributes--- but the underlying mechanism is nevertheless different, and in this case there's no need to be a bias already present in the phenomenon itself, but rather it may be a consequence of the choice of data to be employed. In other words, this is an example of bias \textit{from data to algorithm} in the taxonomy of~\cite{mehrabi2021survey}, i.e. a bias due to data availability, choice and collection. Incidentally, notice that this form of bias ---using a biased proxy of a relevant variable--- might as well happen with the target variable. In this situation, it is the quantity that we need to estimate/predict that is somehow ``flawed''. The fact  that measurement bias depends also on a choice component, which is to say the choice of the dataset, can extend its occurrence in the WYSIWYG worldview as well. Indeed, the choice of ``what to measure'' determines and modifies what is made observable. The eventuality of awareness of biased measurements would probably require mitigation also in WYSIWYG. Alternatively, according to WAE, measurement bias may lie in the mapping between construct and observed space, which is to say between a ``real'' ability of an individual and an observable quantity that tries to measure it.

\begin{figure*}[h]
\centering
\begin{subfigure}[c]{0.5\textwidth}
\centering
\begin{tikzpicture}
    \node[state, fill=gray!50] (r) at (0, 1) {$R$};
    \node[state] (a) at (-1.5, -1) {$A$};
    \node[state, fill=gray!50] (y) at (1.5, 0) {$Y$};
    \path (r) edge (y);
    \path (a) edge (r);
    \path[dashed] (a) edge (y);
\end{tikzpicture}
\caption{Historical bias on feature and target (dashed) variable}
\end{subfigure}%
\begin{subfigure}[c]{0.5\textwidth}
\centering
\begin{tikzpicture}
    \node[state] (r) at (0, 1) {$R$};
    \node[state] (a) at (-1.5, -1) {$A$};
    \node[state, fill=gray!50] (y) at (1.5, 0) {$Y$};
    \node[state, fill=gray!50] (q) at (0, -1) {$Q$};
    \path (r) edge (y);
    \path (q) edge (y);
    \path (a) edge (q);    
\end{tikzpicture}
\caption{Omitted variable bias}
\end{subfigure}
\par\bigskip
\begin{subfigure}[c]{0.5\textwidth}
\centering
\begin{tikzpicture}
    \node[state] (r) at (0, 1) {$R$};
    \node[state] (a) at (-1.5, -1) {$A$};
    \node[state, fill=gray!50] (y) at (1.5, 0) {$Y$};
    \node[state, fill=gray!50] (p) at (0, -1) {$P_R$};
    \path (r) edge (y);
    \path (r) edge (p);
    \path (a) edge (p);    
\end{tikzpicture}
\caption{Measurement bias on $R$}
\end{subfigure}%
\begin{subfigure}[c]{0.5\textwidth}
\centering
\begin{tikzpicture}
    \node[state, fill=gray!50] (r) at (0, 1) {$R$};
    \node[state] (a) at (-1.5, -1) {$A$};
    \node[state] (y) at (1.5, 0) {$Y$};
    \node[state, fill=gray!50] (p) at (1.5, -1) {$P_Y$};
    \path (r) edge (y);
    \path (y) edge (p);
    \path (a) edge (p);    
\end{tikzpicture}
\caption{Measurement bias on $Y$}
\end{subfigure}

\caption{Atomic biases. Grey-filled circles represent variables employed by the model $\hat{f}$.}
\label{fig:causal_graph}
\end{figure*}
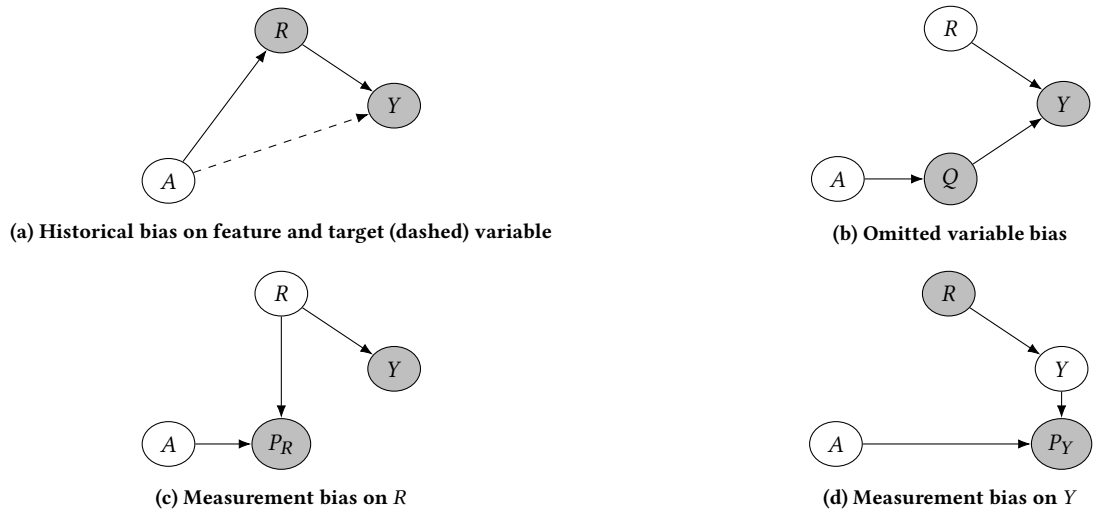

\textbf{Representation bias} occurs when, for some reasons, data are not representative of the world population. 
One subgroup of individuals, e.g. identified by some sensitive characteristic such as ethnicity, age, etc. may be heavily underrepresented. This under-representation may occur in different ways. It may be at random or selecting individuals with disproportionate characteristics with respect to their corresponding world population, e.g. only low-income individuals. In this last case, representation bias may be enough to create disparities in decision making processes based on those data. 
It may be at random, i.e. the subgroup is less numerous than it should be, but without any particular skewness in the other characteristics: in this case this single mechanism is not sufficient to create disparities, but it may exacerbate existing ones. Alternatively, or the under-represented subgroup might contain individuals with disproportionate characteristics with respect to their corresponding world population, e.g. only low-income individuals, or only low-education individuals. In this last case, representation bias may be enough to create disparities in decision making processes based on those data. 
The mechanism underlying the representation disparities should be analyzed on the basis of the assumed worldview and/or chosen EO framework: e.g. if the data has an under-represented ethnic minority one should investigate \emph{why} it is so. If the target population is itself different from world population (i.e it is not merely a poor data collection), then one should consider the reasons by which this ethnic minority is under-represented in the target population, e.g. in the Substantive EO framework one should understand if these reasons have to be regarded as \textit{circumstance} or as consequence of individual \textit{effort}. In the latter case, the representation disparities are not to be considered as ``unfair'' \emph{per se}.

Representation bias is strictly connected to \emph{samplig bias}, in that it embodies problems arising during data collection, e.g. by collecting disproportionately less observations from one subgroup, 
possibly skewed with respect to some characteristics. Like measurement bias, this is a form of bias going \textit{from data to algorithm}.

\textbf{Omitted variable bias} may occur when the collected dataset omits a variable relevant to some specific goal or task. In this case, if the variables that are present in the dataset have some dependence on sensitive characteristics of individuals, then a machine learning model trained on such dataset will learn such dependencies, thus producing outcomes with spurious dependence on sensitive attributes. 
Assuming the Formal EO framework, sensitive features are omitted by default. While this may appear fairer because the decision is made solely on the basis of the relevant attributes, on the other hand it becomes arduous to mitigate on structural biases that affect achievements.
Depending on worldview assumptions or on the chosen EO framework, the mechanism through which the residual variables happen to depend on sensitive individual characteristics should be as well analyzed to understand/decide whether this dependence is legitimate or if they are themselves a consequence of some bias at work.
\vspace{5pt}

The above list of biases should be seen as the set of the most important mechanisms through which ``unfair'' disparities happen to result in data-driven decision making. Notice, however, that in terms of consequences on the data, it may well be that different types of bias result in very similar effects. E.g. representation bias may create in the dataset spurious correlations among sensitive characteristics of individuals and other characteristics relevant to the problem at hand, a situation very similar to the correlations present as a consequence of historical bias.
This reminds us that in reality we are not aware of the type of bias (or biases) affecting the data and that their interpretation depends on former assumptions. Furthermore, starting from specific worldview and framework influence the choices of optimization.

\section{Dataset Generation\label{sec:dataset_generation}}

We propose a simple modeling framework able to simulate the bias-generating mechanisms described in section~\ref{sec:bias_list}.

The rationale behind the model is that of being at the same time sufficiently flexible to accommodate all the main forms of bias generation \emph{while} maintaining a structure as simple and intuitive as possible to  facilitate \textit{Human readability} and ensure \textit{Compactness} avoiding unnecessary complexities that might hide the relevant patterns.  

As noted in section~\ref{sec:bias_list}, following~\cite{mehrabi2021survey} we can distinguish between biases from \emph{users to data} and from \emph{data to algorithm}. Namely, between biases that impact the phenomenon to be studied and thus the dataset, and biases that impact directly the dataset but not the phenomenon itself. Formally, we model the relevant quantities describing a phenomenon as random variables, in particular we label $Y$ the \emph{target} variable, namely the quantity to be estimated or predicted on the basis of other \emph{feature} variables, that we collectively call $X$. As usual, we assume that the underlying phenomenon is described by the formula  
\begin{equation}
    Y = f(X) + \epsilon,
    \label{eq:true_eq}
\end{equation}
where $f$ represents the actual relationship between features and target variables, modulated by some idiosyncratic noise $\epsilon$.

A data-driven decision maker infers from a (training) set of samples $\{(\widetilde{x}_i, y_i)\}_{i=1,\ldots,N}$, an estimate for $f$ that we label $\hat{f}$, thus producing its best estimate for $Y$, namely
\begin{equation}
    \hat{Y} = \hat{f}(\widetilde{X}).
    \label{eq:model_eq}
\end{equation}
The use of $\widetilde{X}$ rather than $X$ describes the fact that the set of variables employed to make inferences about a phenomenon may not coincide with the actual variables that play in that phenomenon. This is precisely what happens in some forms of biases such as measurement bias or omitted variable bias.

Notice that \emph{users-to-data} types of bias impact directly equation~\eqref{eq:true_eq}, while \emph{data-to-algorithm} biases have a role at the level of equation~\eqref{eq:model_eq}.

It is possible to make a schematic representation of building blocks of biases as discussed in section~\ref{sec:bias_list} via Directed Acyclic Graphs (DAGs), representing causal impacts among variables (see e.g.~\cite{pearl2009causality, pearl2018book, peters2014causal}). In general, in order to provide an intuitive grasp on interesting mechanisms and patterns, we shall make reference to the following situation: 
\begin{itemize}
    \item[-] we label with $R$ variables representing \emph{resources} of individuals ---be them economic resources, or personal talents and skills--- which are relevant for the problem, i.e. they directly impact the target $Y$;
    \item[-] we label with $A$ variables indicating sensitive attributes, such as ethnicity, gender, etc.;
    \item[-] we label with $P_R$ proxy variables that we have access to instead of the original variable $R$. E.g. we may think of $R$ as soft skills, or talent, or intelligence, i.e. something that we cannot measure or sometimes even quantify directly;
    \item[-] we label with $Q$ additional variables, that may or may not be relevant for the problem (i.e. impacting $Y$) and that may or may not be impacted either by $R$ or $A$, e.g. the living neighborhood.  
\end{itemize}
With these examples in mind, we can e.g. think of $Y$ ---i.e. the target of our decision making process--- as the repayment of a debt or the work performance. 

In particular, figure~\ref{fig:causal_graph} shows 3 minimal graph representations of historical, omitted variable and measurement biases. Historical bias occurs when the relevant variable $R$ is somehow impacted by sensitive feature $A$. 
Omitted variable bias occurs when, for some reasons, we omit the relevant variable $R$ from our dataset and we employ another variable which happens to be impacted by $A$.
Measurement bias occurs when the relevant variable $R$ is, in general, free of bias, but we cannot access it, thus we employ a proxy $P_R$ which \emph{is impacted} by sensitive characteristic $A$. 
The measurement bias could occur also on the target variable $Y$, when we can access only on a proxy $P_Y$ of the phenomenon that we want to predict.

The following system of equations formalizes the relationships between variables that are used simulate specific form of biases:

\begin{subequations}
\begin{align}
A =& B_A,\quad B_A\sim\mathcal{B}er(p_A);\\
R =& -\b^R_{h} A + N_R,\quad N_R\sim\mathcal{N}(\mu_R, \sigma_R^2);\\
Q =& B_Q,\quad B_Q\ \lvert\ \left(R, A\right) \sim\mathcal{B}in(K, p_Q(R, A)), \\
&p_Q(R, A) = \text{sigmoid}\left(-(\alpha_{RQ}R - \b^Q_{h}A)\right);\nonumber\\
S =& \alpha_R R - \alpha_Q Q - \b^Y_{h} A + N_S,\quad N_S\sim\mathcal{N}(0, \sigma_S^2);\\
Y =& \bm{1}_{\{S > \bar{s}\}}.\\
\end{align}
\label{eq:generator}
\end{subequations}
When simulating measurement bias, either on resources $R$ or on target $Y$, we are going to use the following \emph{proxies} as noisy (and biased) substitutes of the actual variables:
\begin{subequations}
\begin{align}
&P_R = R - \b^R_{m}A + N_{P_R};\quad N_{P_R}\sim\mathcal{N}(0, \sigma_{P_R}^2);\\
&P_S = S - \b^Y_{m} A + N_{P_S};\quad N_{P_S}\sim\mathcal{N}(0, \sigma_{P_S}^2);\\
&P_Y = \bm{1}_{\{P_S > \bar{s}\}}
\end{align}
\label{eq:proxies}
\end{subequations}

By varying the values of the parameters, we are able to generate different aspect of biases as follows: \begin{itemize}
    \item[-] $\b^j_{h}$ determines the presence and amplitude of the \textit{historical bias} on the variable $j\in \{R, Q, Y\}$;
    \item[-] $\b^j_{m}$, when the proxy $P_j$ is used instead of the original variable $j$, governs the intensity of \textit{measurement bias} on $j\in\{R, Y\}$;
    \item[-] $\alpha_R$, $\alpha_Q$ control the linear impact on ($S$ and thus) $Y$ of $R$ and $Q$, respectively; $\alpha_{RQ}$ represents the intensity of the dependence of $Q$ on $R$.
\end{itemize}

\captionsetup[subfigure]{labelformat=empty}
\begin{figure*}
\centering
\begin{subfigure}[c]{0.33\textwidth}
\centering
\begin{tikzpicture}
    \node[state] (r) at (0, 1) {$R$};
    \node[state] (a) at (-1.5, 0) {$A$};
    \node[state] (q) at (0, -1) {$Q$};
    \node[state] (y) at (1.5, 0) {$Y$};
    \path (r) edge (y);
    \path (q) edge (y);
\end{tikzpicture}
\caption{I) No historical bias}
\end{subfigure}%
\begin{subfigure}[c]{0.33\textwidth}
\centering
\begin{tikzpicture}
    \node[state] (r) at (0, 1) {$R$};
    \node[state] (a) at (-1.5, 0) {$A$};
    \node[state] (q) at (0, -1) {$Q$};
    \node[state] (y) at (1.5, 0) {$Y$};
    \path (r) edge (y);
    \path (q) edge (y);
    \path (a) edge (q);    
    \path (a) edge (r);    
\end{tikzpicture}
\caption{II) Historical bias combined}
\end{subfigure}%
\begin{subfigure}[c]{0.33\textwidth}
\centering
\begin{tikzpicture}
    \node[state] (r) at (0, 1) {$R$};
    \node[state] (a) at (-1.5, 0) {$A$};
    \node[state] (q) at (0, -1) {$Q$};
    \node[state] (y) at (1.5, 0) {$Y$};
    \path (r) edge (y);
    \path (a) edge (r);    
    \path (q) edge (y); 
\end{tikzpicture}
\caption{III) Historical bias on $R$}
\end{subfigure}
\par\bigskip
\begin{subfigure}[c]{0.5\textwidth}
\centering
\begin{tikzpicture}
    \node[state] (r) at (0, 1) {$R$};
    \node[state] (a) at (-1.5, 0) {$A$};
    \node[state] (q) at (0, -1) {$Q$};
    \node[state] (y) at (1.5, 0) {$Y$};
    \path (r) edge (y);
    \path (q) edge (y);
    \path (a) edge (y);    
\end{tikzpicture}
\caption{IV) Historical bias on $Y$}
\end{subfigure}%
\begin{subfigure}[c]{0.5\textwidth}
\centering
\begin{tikzpicture}
    \node[state] (r) at (0, 1) {$R$};
    \node[state] (a) at (-1.5, 0) {$A$};
    \node[state] (q) at (0, -1) {$Q$};
    \node[state] (y) at (1.5, 0) {$Y$};
    \path (r) edge (y);
    \path (q) edge (y);
    \path (a) edge (r);
    \path (a) edge (y);
\end{tikzpicture}
\caption{V) Historical bias on $R$ and $Y$}
\end{subfigure}

\caption{The 5 considered scenarios with \emph{users-to-data} biases.}
\label{fig:scenarios}
\end{figure*}
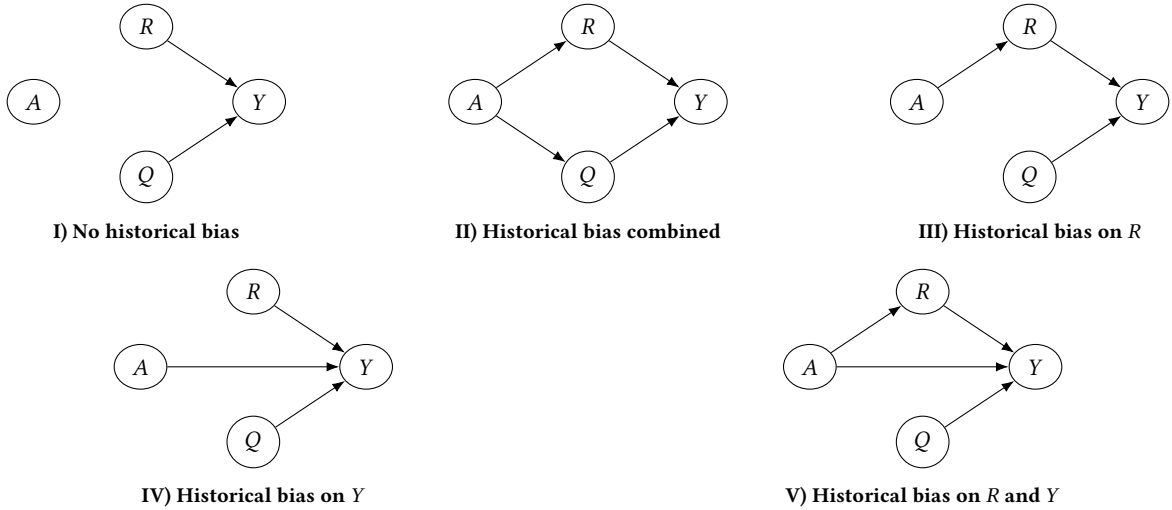

Additionally, in order to account for \emph{representation bias}, we undersample the $A=1$ group. The amount of undersampling is governed by the parameter $p_u$ defined as the proportion of the under-represented group $A=1 $ with respect to the majority group $A=0$. We introduce also the possibility of drawing this undersampling \emph{conditioned} on $R$ by selecting the $A=1$ individuals with lower $R$.
Finally, simulating \emph{omission bias} is as simple as dropping the variable $R$ from the set of features the model uses to estimate $Y$.

\section{Experiments}
We employ our modeling framework to generate a total of twenty-five different datasets by combining a set of five \textit{users-to-data} biases with a set of four \textit{data-to-algorithm} biases, plus a configuration without additional biases. The aim of the experiments is to investigate the effects of these bias combinations on machine learning algorithms.

As clarified in section \ref{sec:bias_list}, \textit{users-to-data} biases impact the phenomenon to be studied and thus the dataset. The five scenarios that we have chosen to explore, represented in figure~\ref{fig:scenarios}, are define as follows:

\begin{itemize}
    \item[I)] \textbf{No historical bias}: 
    \begin{equation}\label{eq:S1}
        Y = f(R,Q) + \epsilon, \quad R,Q \indep A.
    \end{equation}

    \item[II)] \textbf{Historical bias with effect compensation}:
    \begin{equation}\label{eq:S2}
        Y = f(R,Q) + \epsilon,\quad R = R(A),\ Q = Q(A),\ Y \indep A.
    \end{equation}

    \item[III)] \textbf{Historical bias on $R$}: 
    \begin{equation}\label{eq:S3}
        Y = f(R,Q) + \epsilon,\quad R = R(A),\ Q \indep A.
    \end{equation}

    \item[IV)] \textbf{Historical bias on $Y$}: 
    \begin{equation}\label{eq:S4}
        Y = f(R,Q,A) + \epsilon,\quad R,Q \indep A.
    \end{equation}

    \item[V)] \textbf{Historical bias on $R$ and $Y$}: 
    \begin{equation}\label{eq:S5}
        Y = f(R,Q,A) + \epsilon,\quad R=R(A),\  Q \indep A.
    \end{equation}
\end{itemize}

We combine each scenario with the following \textit{data-to-algorithm} biases, having an effect on the variables that we assume we can access in order to train the models:
\begin{itemize}
    \item[-] \textit{no additional bias}; 
    \item[-] \textit{measurement bias} on $R$; 
    \item[-] \textit{omission bias} on $R$; 
    \item[-] \textit{representation bias} on $A=1$; 
    \item[-] \textit{measurement bias} on $Y$. 
\end{itemize}

We thus generate twenty-five synthetic datasets, each of which characterized by a specific choice of the set of parameters discussed in section~\ref{sec:dataset_generation}. Details on the values of the parameters and the complete \texttt{python} code to reproduce the experiments is publicly available at \href{https://github.com/rcrupiISP/ISParity}{\texttt{github.com/rcrupiISP/ISParity}}.

On each generated dataset, we train and test the following three algorithms: (i) a \emph{Random Forest} (RF)~\cite{biau2016random} that aims to maximize performance by utilizing all available variables; (ii) \textit{Blinded Random Forest}~(BRF): a random forest that does not use the sensitive variable $A$ in order to avoid creating decision-making patterns based on it; (iii) \textit{Equalized Random Forest} (ERF): the same model in (i) with a post-processing fairness mitigation block~\cite{bird2020fairlearn} to impose the same rate of $\hat{Y}=1$ between the different classes of $A$ (i.e. imposing Demographic Parity~\cite{barocas-hardt-narayanan}).

Fairness of outcomes is evaluated through \textit{Demographic Parity} difference ($\Delta$DP)~\cite{barocas-hardt-narayanan}, i.e. the difference between the rate of $\hat{Y} = 1$ among the two sensitive groups $A=0$ and $A=1$, namely $\Delta\text{DP} = P(\hat{Y} = 1\ |\ A=0) - P(\hat{Y} = 1\ |\ A=1)$. $\Delta$DP can also be computed for the ``true'' target variable $Y$, thus capturing the dependence of the target variable on the sensitive attribute $A$.
Performance is evaluated by overall \textit{Accuracy} (Acc).
As an additional metric both of fairness and of performance, we compute the difference in accuracy between $A=0$ and $A=1$ groups ($\Delta$Acc).

\begin{table*}
\centering

\caption{\textbf{Experiments results}. Results regarding 25 experiments: 5 \emph{users-to-data} scenarios with varying types of biases (rows) combined with 5 \emph{data-to-algorithm} types of bias. Metrics employed are Accuracy (Acc), Demographic Parity and Accuracy difference between $A=0$ and $A=1$ groups ($\Delta$DP and $\Delta$Acc). $\Delta$DP for the ``true'' target variable $Y$ is also provided for each scenario (values in round brackets).
RF is a random forest trained with all the available information, BRF is a RF blinded with respect to $A$, ERF is a RF with a post-processing fairness mitigation to impose Demographic Parity. Values are expressed as percentage points.}
\label{table:results}

\ra{1.4}
\resizebox{\textwidth}{!}{
\begin{tabular}{@{}l c c c ccc c ccc c ccc c ccc c ccc@{}}
 \toprule
 
 &&&& \multicolumn{19}{c}{\emph{data-to-algorithm} bias} \\
 
 \cmidrule{5-23}
 \emph{users-to-data} bias & \phantom{a} & \phantom{metrics} & \phantom{a} & \multicolumn{3}{c}{no additional bias} & \phantom{a} & \multicolumn{3}{c}{+ meas. bias on $R$} & \phantom{a} & \multicolumn{3}{c}{+ omission bias} & \phantom{a} & \multicolumn{3}{c}{+ representation bias} & \phantom{a} & \multicolumn{3}{c}{+ meas. bias on $Y$}\\
 Scenario &&&&&&&& \multicolumn{3}{c}{($P_R$ substitutes $R$)} && \multicolumn{3}{c}{($R$ is omitted)} & \phantom{ab} & \multicolumn{3}{c}{\makecell[c]{($A=1$ randomly \\undersampled)}} & \phantom{ab} & \multicolumn{3}{c}{\makecell[c]{($P_Y$ substitutes $Y$)}}\\

 \midrule
 && metrics && \makecell[c]{RF} & BRF  & ERF && \makecell[c]{RF} & BRF  & ERF && \makecell[c]{RF} & BRF  & ERF && \makecell[c]{RF} & BRF  & ERF && \makecell[c]{RF} & BRF  & ERF\\ 
 \cmidrule{3-3}\cmidrule{5-7}\cmidrule{9-11}\cmidrule{13-15}\cmidrule{17-19}\cmidrule{21-23}

 \multirow{3}{*}{\makecell[r]{I) No historical\\ bias}} && Acc && 86.4 & 86.4 & 86.4 && 81.2 &	81.1 &	81.2 && 62.8 & 62.8 & 62.8 && 86.3 & 86.3 & 86.3 && 83.3 & 84.5 & 84.3\\
 && $\Delta$DP (0) && 0.1 & 0.1 & 0.0 && 0.8 &	7.4 & 0.5 && 1.2 & 1.2 & 1.0 && 0.7 & 0.6 & 0.4 && 11.2 & 0.1 & 0.0\\
 && $\Delta$Acc  && 0.1 & 0.1 & 0.1 && 0.5 & 0.7 & 0.5 && 0.4 & 0.4 & 0.4 && 2.0 & 1.8 & 2.0 && 4.0 & 0.1 & 0.0\\

 \cmidrule{5-23}
 \multirow{3}{*}{\makecell[r]{II) Historical bias \\ compensation}} && Acc && 86.3 & 86.4 & 86.4 && 81.1 & 81.1 & 81.1 && 62.7 & 61.6 & 62.4 && 86.3 & 86.3 & 86.4 && 83.7 & 84.2 & 84.6\\
 && $\Delta$DP (0.9) && 0.1 & 0.8 & 0.3 && 2.2 & 5.6 & 0.9 && 9.8 & 27.1 & 0.9 && 1.6 & 0.7 & 0.6 && 11.5 & 3.9 & 0.4\\
 && $\Delta$Acc && 0.0 & 0.1 & 0.0 && 0.4 & 0.7 & 0.3 && 0.7 & 1.6 & 0.0 && 1.4 & 1.5 & 1.2 && 3.4 & 0.7 & 0.5 \\

 \cmidrule{5-23}
 \multirow{3}{*}{\makecell[r]{III) Historical\\ bias on $R$}} && Acc && 86.7 & 86.6 & 85.5 && 81.5 &	81.4 &	80.6 && 64.5 & 62.6 & 62.6 && 86.4 & 86.3 & 86.2 && 84.0 & 84.9 & 83.7\\
 && $\Delta$DP (5) && 12.6 & 12.9 & 0.0 && 11.8 & 19.1 & 0.7 && 36.7 & 1.2 & 1.0 && 11.2 & 10.3 & 0.0 && 25.1 & 13.1 & 0.1\\
 && $\Delta$Acc && 2.2 & 2.1 & 2.3 && 2.5 & 2.2 & 2.3 && 1.2 & 2.6 & 2.6 && 0.9 & 1.0 & 1.3  && 1.7 & 1.5 & 4.9\\

 \cmidrule{5-23}
 \multirow{3}{*}{\makecell[r]{IV) Historical\\ bias on $Y$}} && Acc && 86.5 & 85.5 & 85.5 && 81.5	& 81.3 & 80.6 && 64.5 & 62.6 & 62.6 && 86.4 & 86.1 & 86.2 && 84.3 & 83.5 & 83.9\\
 && $\Delta$DP (12) && 11.2 & 0.1 & 0.0 && 10.9 &	7.4 &	0.6 && 36.7 & 1.2 & 1.0 && 10.7 & 0.5 & 0.0 && 23.1 & 0.2 & 0.1\\  
 && $\Delta$Acc && 1.7 & 0.0 & 2.4 && 2.7 &	2.9 &	2.2 && 1.3 & 2.6 & 2.6 && 1.0 & 1.6 & 1.3  && 1.1 & 5.7 & 4.7\\

 \cmidrule{5-23}
 \multirow{3}{*}{\makecell[r]{V) Historical bias\\ on $R$ \emph{and} $Y$}} && Acc && 86.9 & 86.0 & 83.1 && 82.0 &	81.6 &	78.7 && 66.4 & 62.1 & 62.0 && 86.5 & 86.3 & 85.6 && 84.6 & 84.3 & 81.2\\
 && $\Delta$DP (7) && 25.1 & 13.1 & 0.1 && 27.0 &	18.8 &	0.6 && 36.7 & 1.2 & 1.0 && 19.1 & 10.3 & 1.3 && 37.4 & 13.3 & 0.1\\
 && $\Delta$Acc && 4.0 & 3.8 & 3.7 && 5.1 &	5.0 & 4.6 && 4.7 & 4.1 & 4.1 && 4.2 & 1.4 & 6.5 && 0.1 & 6.6 & 11.3\\

 \bottomrule
\end{tabular}
}

\end{table*}

\section{Discussion}

Table~\ref{table:results} summarizes the results obtained by the twenty-five experiments on generated datasets, on the basis of which we draw the following non-exhaustive list of insights:

\textbf{On the effect of equalization}: as expected, the adoption of a model to enforce a group equalization of the outcomes results in lower performance in the majority of scenarios considered. This gap is proportional to the level of disparity present in the data. On the contrary, when there is representation bias, the effects on performance are less visible, since enforcing equalization impacts fewer instances. This suggests that observing significant reductions in $\Delta$DP with negligible performance deterioration is not \emph{per se} sufficient to conclude that a mitigation strategy has no effects on performance ---as expressed by numerous studies in the field of fairness--- since the protected class happens often to be a minority. Observing the difference in accuracy ($\Delta$Acc), on the other hand, is a much more robust indicator because it is unaffected by the different numerosity of the protected subgroups.

\textbf{On the effect of blindness}: excluding sensitive attributes when training machine learning models ---as suggested by the Formal~EO framework--- has a negative impact on performance only when $Y$ is directly influenced by $A$. When the factors that determine $Y$ contain historical bias, this technique has no benefit in terms of demographic parity. It's worth noting that in the presence of measurement bias in scenarios I, II, and III, the effects on $\Delta$DP are quite worse than when $A$ is included in the model. This occurs because the machine learning model actually needs $A$ to remove the excess of dependence induced by the inclusion of the proxy.

\textbf{On the performance-bias trade-off in moral frameworks}: one crucial aspect to keep in mind when developing a decision system is that what is and what is not a \emph{bias} depends on the moral framework and worldview chosen in the first place. In this regard, the employment of a modeling strategy over another (e.g. using BRF instead of ERF) is in response to the chosen worldview. Thus, \emph{given a specific moral view}, there are circumstances where it is justifiable to prefer performance over group equality and vice versa. Notice that the choice of a framework needs often be complemented with additional (crucial) decisions, e.g. choosing Substantive EO \emph{per se} is not enough to understand whether some phenomenon underlying data is or not a bias, since one needs to agree on which characteristics are  to be considered as ``individual effort'' (as opposed to ``circumstance''), and thus legitimately employed as rationale of decisions.

\textbf{On the interpretation of results}: in this work, we simulate the presence of one or more biases and observe the corresponding effects on data. In real-world scenarios, data are typically observed without knowing the exact bias-generation mechanism underlying them. It is the assumption on the data generation mechanism that shapes the interpretation of biases present in the context under consideration.

This concept becomes more evident when the target variable in the data serves as proxy for the true target. In our experiments with measurement bias on $Y$, we trained the models with the proxy and evaluated the results with the real variable. This approach, even if unlikely in real-world situations, demonstrates that reducing bias can also lead to improved performances. These set of experiments shows that under a specific worldview, mitigation strategies that result in lower performance on the target variable in the observed space, may lead to higher performance on the ``true'' target.

\textbf{On the absence of bias}: when operating in an ideal bias-free environment, i.e. when any possible relevant characteristic is independent of sensitive features, all machine learning strategies provide the same performance and fairness results. Furthermore, in this ideal situation, all worldviews would agree, even if they are mutually incompatible in the general case. Regardless of how trivial this conclusion may be, it is essential to mention it in order to promote any legal, social, or strategic initiative with the goal of eliminating society biases.


\section{Conclusion}

In this work we have investigated how different types of bias impact fairness and performance metrics of machine learning models by introducing of a unified framework for generating synthetic data. Regardless of whether the test scenarios used to draw conclusions are synthetic, we have discussed moral frameworks and worldviews that can be applied in real-world situations.

We release the presented model framework as open-source at \href{https://github.com/rcrupiISP/ISParity}{\texttt{github.com/rcrupiISP/ISParity}} in order to encourage further research into the challenge of bias in data. 
For example, one crucial aspect to be further analyzed is the sensitiveness of results with varying magnitude of parameters. Another is to control the effect of random fluctuations by averaging results of many simulations at once for each choice of the parameters set. 


\bibliographystyle{ACM-Reference-Format}
\bibliography{references.bib}


\begin{thebibliography}{33}


\ifx \showCODEN    \undefined \def \showCODEN     #1{\unskip}     \fi
\ifx \showDOI      \undefined \def \showDOI       #1{#1}\fi
\ifx \showISBNx    \undefined \def \showISBNx     #1{\unskip}     \fi
\ifx \showISBNxiii \undefined \def \showISBNxiii  #1{\unskip}     \fi
\ifx \showISSN     \undefined \def \showISSN      #1{\unskip}     \fi
\ifx \showLCCN     \undefined \def \showLCCN      #1{\unskip}     \fi
\ifx \shownote     \undefined \def \shownote      #1{#1}          \fi
\ifx \showarticletitle \undefined \def \showarticletitle #1{#1}   \fi
\ifx \showURL      \undefined \def \showURL       {\relax}        \fi
\providecommand\bibfield[2]{#2}
\providecommand\bibinfo[2]{#2}
\providecommand\natexlab[1]{#1}
\providecommand\showeprint[2][]{arXiv:#2}

\bibitem[Agarwal et~al\mbox{.}(2018)]%
        {agarwal2018reductions}
\bibfield{author}{\bibinfo{person}{Alekh Agarwal}, \bibinfo{person}{Alina
  Beygelzimer}, \bibinfo{person}{Miroslav Dud{\'\i}k}, \bibinfo{person}{John
  Langford}, {and} \bibinfo{person}{Hanna Wallach}.}
  \bibinfo{year}{2018}\natexlab{}.
\newblock \showarticletitle{A reductions approach to fair classification}. In
  \bibinfo{booktitle}{\emph{International Conference on Machine Learning}}.
  PMLR, \bibinfo{pages}{60--69}.
\newblock


\bibitem[Aggarwal and Philip(2008)]%
        {aggarwal2008privacy}
\bibfield{author}{\bibinfo{person}{Charu~C Aggarwal} {and}
  \bibinfo{person}{S~Yu Philip}.} \bibinfo{year}{2008}\natexlab{}.
\newblock \bibinfo{booktitle}{\emph{Privacy-preserving data mining: models and
  algorithms}}.
\newblock \bibinfo{publisher}{Springer Science \& Business Media}.
\newblock


\bibitem[Angwin et~al\mbox{.}(2016)]%
        {angwin2016machine}
\bibfield{author}{\bibinfo{person}{Julia Angwin}, \bibinfo{person}{Jeff
  Larson}, \bibinfo{person}{Surya Mattu}, {and} \bibinfo{person}{Lauren
  Kirchner}.} \bibinfo{year}{2016}\natexlab{}.
\newblock \showarticletitle{Machine bias: There’s software used across the
  country to predict future criminals, and it’s biased against blacks}.
\newblock \bibinfo{journal}{\emph{ProPublica}} (\bibinfo{year}{2016}).
\newblock


\bibitem[Assefa et~al\mbox{.}(2020)]%
        {assefa2020generating}
\bibfield{author}{\bibinfo{person}{Samuel~A Assefa}, \bibinfo{person}{Danial
  Dervovic}, \bibinfo{person}{Mahmoud Mahfouz}, \bibinfo{person}{Robert~E
  Tillman}, \bibinfo{person}{Prashant Reddy}, {and} \bibinfo{person}{Manuela
  Veloso}.} \bibinfo{year}{2020}\natexlab{}.
\newblock \showarticletitle{Generating synthetic data in finance:
  opportunities, challenges and pitfalls}. In
  \bibinfo{booktitle}{\emph{Proceedings of the First ACM International
  Conference on AI in Finance}}. \bibinfo{pages}{1--8}.
\newblock


\bibitem[Barocas et~al\mbox{.}(2019)]%
        {barocas-hardt-narayanan}
\bibfield{author}{\bibinfo{person}{Solon Barocas}, \bibinfo{person}{Moritz
  Hardt}, {and} \bibinfo{person}{Arvind Narayanan}.}
  \bibinfo{year}{2019}\natexlab{}.
\newblock \bibinfo{booktitle}{\emph{Fairness and Machine Learning}}.
\newblock \bibinfo{publisher}{fairmlbook.org}.
\newblock
\newblock
\shownote{\url{http://www.fairmlbook.org}}.


\bibitem[Biau and Scornet(2016)]%
        {biau2016random}
\bibfield{author}{\bibinfo{person}{G{\'e}rard Biau} {and}
  \bibinfo{person}{Erwan Scornet}.} \bibinfo{year}{2016}\natexlab{}.
\newblock \showarticletitle{A random forest guided tour}.
\newblock \bibinfo{journal}{\emph{Test}} \bibinfo{volume}{25},
  \bibinfo{number}{2} (\bibinfo{year}{2016}), \bibinfo{pages}{197--227}.
\newblock


\bibitem[Bird et~al\mbox{.}(2020)]%
        {bird2020fairlearn}
\bibfield{author}{\bibinfo{person}{Sarah Bird}, \bibinfo{person}{Miro
  Dud{\'i}k}, \bibinfo{person}{Richard Edgar}, \bibinfo{person}{Brandon Horn},
  \bibinfo{person}{Roman Lutz}, \bibinfo{person}{Vanessa Milan},
  \bibinfo{person}{Mehrnoosh Sameki}, \bibinfo{person}{Hanna Wallach}, {and}
  \bibinfo{person}{Kathleen Walker}.} \bibinfo{year}{2020}\natexlab{}.
\newblock \bibinfo{booktitle}{\emph{Fairlearn: A toolkit for assessing and
  improving fairness in {AI}}}.
\newblock \bibinfo{type}{{T}echnical {R}eport} MSR-TR-2020-32.
  \bibinfo{institution}{Microsoft}.
\newblock
\urldef\tempurl%
\url{https://www.microsoft.com/en-us/research/publication/fairlearn-a-toolkit-for-assessing-and-improving-fairness-in-ai/}
\showURL{%
\tempurl}


\bibitem[Castelnovo et~al\mbox{.}(2020)]%
        {castelnovo2020befair}
\bibfield{author}{\bibinfo{person}{Alessandro Castelnovo},
  \bibinfo{person}{Riccardo Crupi}, \bibinfo{person}{Giulia Del~Gamba},
  \bibinfo{person}{Greta Greco}, \bibinfo{person}{Aisha Naseer},
  \bibinfo{person}{Daniele Regoli}, {and} \bibinfo{person}{Beatriz San~Miguel
  Gonzalez}.} \bibinfo{year}{2020}\natexlab{}.
\newblock \showarticletitle{BeFair: Addressing Fairness in the Banking Sector}.
  In \bibinfo{booktitle}{\emph{2020 IEEE International Conference on Big Data
  (Big Data)}}. IEEE, \bibinfo{pages}{3652--3661}.
\newblock


\bibitem[Castelnovo et~al\mbox{.}(2022)]%
        {castelnovo2022clarification}
\bibfield{author}{\bibinfo{person}{Alessandro Castelnovo},
  \bibinfo{person}{Riccardo Crupi}, \bibinfo{person}{Greta Greco},
  \bibinfo{person}{Daniele Regoli}, \bibinfo{person}{Ilaria~Giuseppina Penco},
  {and} \bibinfo{person}{Andrea~Claudio Cosentini}.}
  \bibinfo{year}{2022}\natexlab{}.
\newblock \showarticletitle{A clarification of the nuances in the fairness
  metrics landscape}.
\newblock \bibinfo{journal}{\emph{Scientific Reports}} \bibinfo{volume}{12},
  \bibinfo{number}{1} (\bibinfo{year}{2022}), \bibinfo{pages}{1--21}.
\newblock


\bibitem[Castelnovo et~al\mbox{.}(2021)]%
        {castelnovo2021towards}
\bibfield{author}{\bibinfo{person}{Alessandro Castelnovo},
  \bibinfo{person}{Lorenzo Malandri}, \bibinfo{person}{Fabio Mercorio},
  \bibinfo{person}{Mario Mezzanzanica}, {and} \bibinfo{person}{Andrea
  Cosentini}.} \bibinfo{year}{2021}\natexlab{}.
\newblock \showarticletitle{Towards Fairness Through Time}. In
  \bibinfo{booktitle}{\emph{Joint European Conference on Machine Learning and
  Knowledge Discovery in Databases}}. Springer, \bibinfo{pages}{647--663}.
\newblock


\bibitem[D'Amour et~al\mbox{.}(2020)]%
        {d2020fairness}
\bibfield{author}{\bibinfo{person}{Alexander D'Amour}, \bibinfo{person}{Hansa
  Srinivasan}, \bibinfo{person}{James Atwood}, \bibinfo{person}{Pallavi
  Baljekar}, \bibinfo{person}{David Sculley}, {and} \bibinfo{person}{Yoni
  Halpern}.} \bibinfo{year}{2020}\natexlab{}.
\newblock \showarticletitle{Fairness is not static: deeper understanding of
  long term fairness via simulation studies}. In
  \bibinfo{booktitle}{\emph{Proceedings of the 2020 Conference on Fairness,
  Accountability, and Transparency}}. \bibinfo{pages}{525--534}.
\newblock


\bibitem[Dwork et~al\mbox{.}(2012)]%
        {dwork2012fairness}
\bibfield{author}{\bibinfo{person}{Cynthia Dwork}, \bibinfo{person}{Moritz
  Hardt}, \bibinfo{person}{Toniann Pitassi}, \bibinfo{person}{Omer Reingold},
  {and} \bibinfo{person}{Richard Zemel}.} \bibinfo{year}{2012}\natexlab{}.
\newblock \showarticletitle{Fairness through awareness}. In
  \bibinfo{booktitle}{\emph{Proceedings of the 3rd innovations in theoretical
  computer science conference}}. \bibinfo{pages}{214--226}.
\newblock


\bibitem[Fleurbaey(2008)]%
        {fleurbaey2008fairness}
\bibfield{author}{\bibinfo{person}{Marc Fleurbaey}.}
  \bibinfo{year}{2008}\natexlab{}.
\newblock \bibinfo{booktitle}{\emph{Fairness, responsibility, and welfare}}.
\newblock \bibinfo{publisher}{OUP Oxford}.
\newblock


\bibitem[Friedler et~al\mbox{.}(2021)]%
        {friedler2021possibility}
\bibfield{author}{\bibinfo{person}{Sorelle~A Friedler}, \bibinfo{person}{Carlos
  Scheidegger}, {and} \bibinfo{person}{Suresh Venkatasubramanian}.}
  \bibinfo{year}{2021}\natexlab{}.
\newblock \showarticletitle{The (im) possibility of fairness: Different value
  systems require different mechanisms for fair decision making}.
\newblock \bibinfo{journal}{\emph{Commun. ACM}} \bibinfo{volume}{64},
  \bibinfo{number}{4} (\bibinfo{year}{2021}), \bibinfo{pages}{136--143}.
\newblock


\bibitem[Hardt et~al\mbox{.}(2016)]%
        {hardt2016equality}
\bibfield{author}{\bibinfo{person}{Moritz Hardt}, \bibinfo{person}{Eric Price},
  {and} \bibinfo{person}{Nati Srebro}.} \bibinfo{year}{2016}\natexlab{}.
\newblock \showarticletitle{Equality of opportunity in supervised learning}. In
  \bibinfo{booktitle}{\emph{Advances in neural information processing
  systems}}. \bibinfo{pages}{3315--3323}.
\newblock


\bibitem[Heidari et~al\mbox{.}(2019)]%
        {heidari2019moral}
\bibfield{author}{\bibinfo{person}{Hoda Heidari}, \bibinfo{person}{Michele
  Loi}, \bibinfo{person}{Krishna~P Gummadi}, {and} \bibinfo{person}{Andreas
  Krause}.} \bibinfo{year}{2019}\natexlab{}.
\newblock \showarticletitle{A moral framework for understanding fair ml through
  economic models of equality of opportunity}. In
  \bibinfo{booktitle}{\emph{Proceedings of the conference on fairness,
  accountability, and transparency}}. \bibinfo{pages}{181--190}.
\newblock


\bibitem[Hertweck et~al\mbox{.}(2021)]%
        {hertweck2021moral}
\bibfield{author}{\bibinfo{person}{Corinna Hertweck},
  \bibinfo{person}{Christoph Heitz}, {and} \bibinfo{person}{Michele Loi}.}
  \bibinfo{year}{2021}\natexlab{}.
\newblock \showarticletitle{On the moral justification of statistical parity}.
  In \bibinfo{booktitle}{\emph{Proceedings of the 2021 ACM Conference on
  Fairness, Accountability, and Transparency}}. \bibinfo{pages}{747--757}.
\newblock


\bibitem[Kamiran and Calders(2009)]%
        {kamiran2009classifying}
\bibfield{author}{\bibinfo{person}{Faisal Kamiran} {and} \bibinfo{person}{Toon
  Calders}.} \bibinfo{year}{2009}\natexlab{}.
\newblock \showarticletitle{Classifying without discriminating}. In
  \bibinfo{booktitle}{\emph{2009 2nd International Conference on Computer,
  Control and Communication}}. IEEE, \bibinfo{pages}{1--6}.
\newblock


\bibitem[Le~Quy et~al\mbox{.}(2022)]%
        {le2022survey}
\bibfield{author}{\bibinfo{person}{Tai Le~Quy}, \bibinfo{person}{Arjun Roy},
  \bibinfo{person}{Vasileios Iosifidis}, \bibinfo{person}{Wenbin Zhang}, {and}
  \bibinfo{person}{Eirini Ntoutsi}.} \bibinfo{year}{2022}\natexlab{}.
\newblock \showarticletitle{A survey on datasets for fairness-aware machine
  learning}.
\newblock \bibinfo{journal}{\emph{Wiley Interdisciplinary Reviews: Data Mining
  and Knowledge Discovery}} (\bibinfo{year}{2022}), \bibinfo{pages}{e1452}.
\newblock


\bibitem[Loh et~al\mbox{.}(2019)]%
        {loh2019subgroup}
\bibfield{author}{\bibinfo{person}{Wei-Yin Loh}, \bibinfo{person}{Luxi Cao},
  {and} \bibinfo{person}{Peigen Zhou}.} \bibinfo{year}{2019}\natexlab{}.
\newblock \showarticletitle{Subgroup identification for precision medicine: A
  comparative review of 13 methods}.
\newblock \bibinfo{journal}{\emph{Wiley Interdisciplinary Reviews: Data Mining
  and Knowledge Discovery}} \bibinfo{volume}{9}, \bibinfo{number}{5}
  (\bibinfo{year}{2019}), \bibinfo{pages}{e1326}.
\newblock


\bibitem[Lohia et~al\mbox{.}(2019)]%
        {lohia2019bias}
\bibfield{author}{\bibinfo{person}{Pranay~K Lohia},
  \bibinfo{person}{Karthikeyan~Natesan Ramamurthy}, \bibinfo{person}{Manish
  Bhide}, \bibinfo{person}{Diptikalyan Saha}, \bibinfo{person}{Kush~R
  Varshney}, {and} \bibinfo{person}{Ruchir Puri}.}
  \bibinfo{year}{2019}\natexlab{}.
\newblock \showarticletitle{Bias mitigation post-processing for individual and
  group fairness}. In \bibinfo{booktitle}{\emph{Icassp 2019-2019 ieee
  international conference on acoustics, speech and signal processing
  (icassp)}}. IEEE, \bibinfo{pages}{2847--2851}.
\newblock


\bibitem[Madaio et~al\mbox{.}(2020)]%
        {madaio2020co}
\bibfield{author}{\bibinfo{person}{Michael~A Madaio}, \bibinfo{person}{Luke
  Stark}, \bibinfo{person}{Jennifer Wortman~Vaughan}, {and}
  \bibinfo{person}{Hanna Wallach}.} \bibinfo{year}{2020}\natexlab{}.
\newblock \showarticletitle{Co-designing checklists to understand
  organizational challenges and opportunities around fairness in ai}. In
  \bibinfo{booktitle}{\emph{Proceedings of the 2020 CHI Conference on Human
  Factors in Computing Systems}}. \bibinfo{pages}{1--14}.
\newblock


\bibitem[Majeed and Lee(2020)]%
        {majeed2020anonymization}
\bibfield{author}{\bibinfo{person}{Abdul Majeed} {and}
  \bibinfo{person}{Sungchang Lee}.} \bibinfo{year}{2020}\natexlab{}.
\newblock \showarticletitle{Anonymization techniques for privacy preserving
  data publishing: A comprehensive survey}.
\newblock \bibinfo{journal}{\emph{IEEE access}}  \bibinfo{volume}{9}
  (\bibinfo{year}{2020}), \bibinfo{pages}{8512--8545}.
\newblock


\bibitem[Mehrabi et~al\mbox{.}(2021)]%
        {mehrabi2021survey}
\bibfield{author}{\bibinfo{person}{Ninareh Mehrabi}, \bibinfo{person}{Fred
  Morstatter}, \bibinfo{person}{Nripsuta Saxena}, \bibinfo{person}{Kristina
  Lerman}, {and} \bibinfo{person}{Aram Galstyan}.}
  \bibinfo{year}{2021}\natexlab{}.
\newblock \showarticletitle{A survey on bias and fairness in machine learning}.
\newblock \bibinfo{journal}{\emph{ACM Computing Surveys (CSUR)}}
  \bibinfo{volume}{54}, \bibinfo{number}{6} (\bibinfo{year}{2021}),
  \bibinfo{pages}{1--35}.
\newblock


\bibitem[Ntoutsi et~al\mbox{.}(2020)]%
        {ntoutsi2020bias}
\bibfield{author}{\bibinfo{person}{Eirini Ntoutsi}, \bibinfo{person}{Pavlos
  Fafalios}, \bibinfo{person}{Ujwal Gadiraju}, \bibinfo{person}{Vasileios
  Iosifidis}, \bibinfo{person}{Wolfgang Nejdl}, \bibinfo{person}{Maria-Esther
  Vidal}, \bibinfo{person}{Salvatore Ruggieri}, \bibinfo{person}{Franco
  Turini}, \bibinfo{person}{Symeon Papadopoulos}, \bibinfo{person}{Emmanouil
  Krasanakis}, {et~al\mbox{.}}} \bibinfo{year}{2020}\natexlab{}.
\newblock \showarticletitle{Bias in data-driven artificial intelligence
  systems—An introductory survey}.
\newblock \bibinfo{journal}{\emph{Wiley Interdisciplinary Reviews: Data Mining
  and Knowledge Discovery}} \bibinfo{volume}{10}, \bibinfo{number}{3}
  (\bibinfo{year}{2020}), \bibinfo{pages}{e1356}.
\newblock


\bibitem[Pearl(2009)]%
        {pearl2009causality}
\bibfield{author}{\bibinfo{person}{Judea Pearl}.}
  \bibinfo{year}{2009}\natexlab{}.
\newblock \bibinfo{booktitle}{\emph{Causality}}.
\newblock \bibinfo{publisher}{Cambridge university press}.
\newblock


\bibitem[Pearl and Mackenzie(2018)]%
        {pearl2018book}
\bibfield{author}{\bibinfo{person}{Judea Pearl} {and} \bibinfo{person}{Dana
  Mackenzie}.} \bibinfo{year}{2018}\natexlab{}.
\newblock \bibinfo{booktitle}{\emph{The book of why: the new science of cause
  and effect}}.
\newblock \bibinfo{publisher}{Basic Books}.
\newblock


\bibitem[Peters et~al\mbox{.}(2014)]%
        {peters2014causal}
\bibfield{author}{\bibinfo{person}{Jonas Peters}, \bibinfo{person}{Joris~M
  Mooij}, \bibinfo{person}{Dominik Janzing}, {and} \bibinfo{person}{Bernhard
  Sch{\"o}lkopf}.} \bibinfo{year}{2014}\natexlab{}.
\newblock \showarticletitle{Causal discovery with continuous additive noise
  models}.
\newblock \bibinfo{journal}{\emph{Journal of Machine Learning Research}}
  \bibinfo{volume}{15}, \bibinfo{number}{58} (\bibinfo{year}{2014}).
\newblock


\bibitem[Raghunathan(2021)]%
        {raghunathan2021synthetic}
\bibfield{author}{\bibinfo{person}{Trivellore~E Raghunathan}.}
  \bibinfo{year}{2021}\natexlab{}.
\newblock \showarticletitle{Synthetic data}.
\newblock \bibinfo{journal}{\emph{Annual Review of Statistics and Its
  Application}}  \bibinfo{volume}{8} (\bibinfo{year}{2021}),
  \bibinfo{pages}{129--140}.
\newblock


\bibitem[Rawls(2004)]%
        {rawls2004theory}
\bibfield{author}{\bibinfo{person}{John Rawls}.}
  \bibinfo{year}{2004}\natexlab{}.
\newblock \showarticletitle{A theory of justice}.
\newblock In \bibinfo{booktitle}{\emph{Ethics}}.
  \bibinfo{publisher}{Routledge}, \bibinfo{pages}{229--234}.
\newblock


\bibitem[Reddy et~al\mbox{.}(2021)]%
        {reddy2021benchmarking}
\bibfield{author}{\bibinfo{person}{Charan Reddy}, \bibinfo{person}{Deepak
  Sharma}, \bibinfo{person}{Soroush Mehri}, \bibinfo{person}{Adriana
  Romero-Soriano}, \bibinfo{person}{Samira Shabanian}, {and}
  \bibinfo{person}{Sina Honari}.} \bibinfo{year}{2021}\natexlab{}.
\newblock \showarticletitle{Benchmarking bias mitigation algorithms in
  representation learning through fairness metrics}. In
  \bibinfo{booktitle}{\emph{Thirty-fifth Conference on Neural Information
  Processing Systems Datasets and Benchmarks Track (Round 1)}}.
\newblock


\bibitem[Roemer and Trannoy(2015)]%
        {roemer2015equality}
\bibfield{author}{\bibinfo{person}{John~E Roemer} {and} \bibinfo{person}{Alain
  Trannoy}.} \bibinfo{year}{2015}\natexlab{}.
\newblock \showarticletitle{Equality of opportunity}.
\newblock In \bibinfo{booktitle}{\emph{Handbook of income distribution}}.
  Vol.~\bibinfo{volume}{2}. \bibinfo{publisher}{Elsevier},
  \bibinfo{pages}{217--300}.
\newblock


\bibitem[Surendra and Mohan(2017)]%
        {surendra2017review}
\bibfield{author}{\bibinfo{person}{HMHS Surendra} {and} \bibinfo{person}{HS
  Mohan}.} \bibinfo{year}{2017}\natexlab{}.
\newblock \showarticletitle{A review of synthetic data generation methods for
  privacy preserving data publishing}.
\newblock \bibinfo{journal}{\emph{International Journal of Scientific \&
  Technology Research}} \bibinfo{volume}{6}, \bibinfo{number}{3}
  (\bibinfo{year}{2017}), \bibinfo{pages}{95--101}.
\newblock


\end{thebibliography}


\end{document}